\documentclass[manuscript,screen,nonacm,sigconf]{acmart}

\setcopyright{none}
\renewcommand\footnotetextcopyrightpermission[1]{}

\usepackage{amsmath,amsfonts,bm}

\def\eqref#1{equation~\ref{#1}}

\def\1{\bm{1}}

\DeclareMathAlphabet{\mathsfit}{\encodingdefault}{\sfdefault}{m}{sl}
\SetMathAlphabet{\mathsfit}{bold}{\encodingdefault}{\sfdefault}{bx}{n}

\usepackage{amsmath,amsfonts}
\usepackage{booktabs}
\usepackage{multirow}
\usepackage{graphicx}
\usepackage{algorithm}
\usepackage{float}
\usepackage{algpseudocode}
\usepackage{longtable}
\usepackage{tikz,pgfplots}
\usepackage{placeins}
\pgfplotsset{compat=1.18}
\usetikzlibrary{arrows.meta,positioning,calc,fit,shapes.geometric,decorations.pathreplacing}

\title{Operator Boosting Produces Pareto-Efficient PDE Surrogates}

\author{Lennon J. Shikhman}
\correspondingauthor
\email{lj@shikhman.net}
\affiliation{%
  \institution{Georgia Institute of Technology}
  \department{College of Computing}
  \city{Atlanta}
  \state{Georgia}
  \country{USA}}

\begin{document}

\begin{abstract}
Neural operators are widely used as surrogate solution maps for partial differential equations (PDEs), but full-size models can be costly to store, deploy, and evaluate in many-query scientific workflows. This work introduces \emph{Operator Boosting}, a stagewise residual-learning framework for constructing compact neural-operator surrogates directly, rather than training a large model and compressing it afterward. Starting from the empirical mean predictor in normalized output coordinates, the method trains a sequence of tiny same-family neural operators on residual fields and incorporates each correction through validation-selected shrinkage. We instantiate the framework with Fourier neural operators (FNOs), DeepONets, and convolutional neural operators (CNOs), and compare boosted tiny stacks against full-size monolithic baselines across one-, two-, and three-dimensional PDE benchmarks from PDEBench, APEBench, and The Well. Across 30 dataset--architecture pairs, 21 show positive mean accuracy gains and 17 have positive confidence intervals, while all boosted stacks reduce trainable parameter count by approximately 72--95\%. Best-model comparisons show empirical Pareto improvements on 7 of 10 completed PDE benchmarks, including two-dimensional Navier--Stokes, shallow-water dynamics, Darcy flow, one-dimensional transport and reaction systems, and three-dimensional compressible Navier--Stokes. These results show that Operator Boosting often improves the empirical accuracy--parameter Pareto frontier of neural PDE surrogates, while also exposing PDE- and architecture-dependent regimes where residual boosting fails to offset compression.
\end{abstract}


\begin{teaserfigure}
  \includegraphics[width=\textwidth]{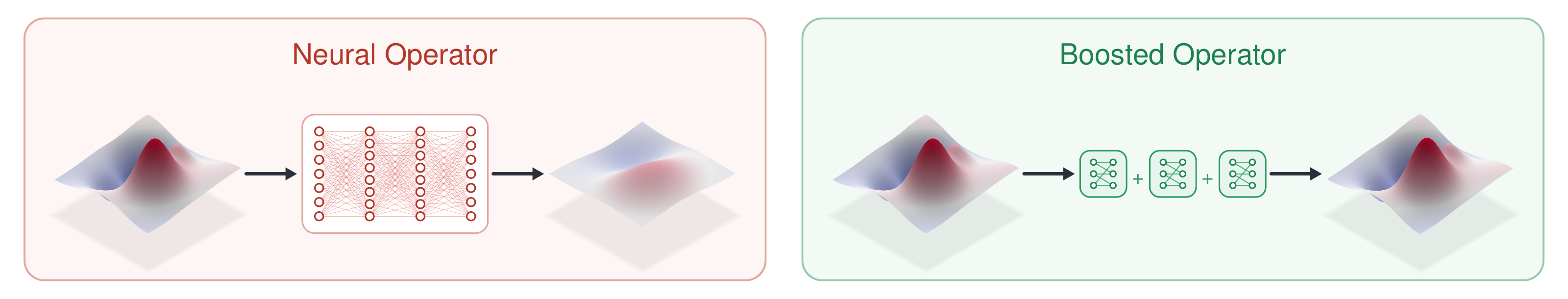}
  \caption{\textbf{Operator Boosting.} A full-size monolithic neural
  operator (left) is replaced by a compact additive stack of tiny
  same-family operators trained stagewise on residual fields (right).
  Across 30 dataset--architecture pairs the boosted stacks reduce
  trainable parameters by 72--95\%, and the best compact stack
  empirically Pareto-dominates the best full-size baseline on 7 of 10
  PDE benchmarks, matching or improving accuracy at a fraction of the
  model size.}
  \Description{A two-panel schematic comparing two neural-operator
  surrogates. The left panel, titled ``Monolithic Operator,'' shows a
  three-dimensional input field surface passing through a single large,
  densely connected neural network and producing an output surface that
  is visibly flattened and blurred relative to the input. The right
  panel, titled ``Operator Boosting,'' shows the same input field
  passing through the sum of three small neural networks and producing
  an output surface that closely matches the input. The left network is
  large and the three right-hand networks are small, conveying far fewer
  parameters; the closer match on the right conveys higher accuracy.}
  \label{fig:teaser}
\end{teaserfigure}


\maketitle

\section{Introduction}

Partial differential equations (PDEs) are central to scientific computing, but repeated high-fidelity simulation can be prohibitively expensive in many-query settings such as design optimization, uncertainty quantification, inverse problems, control, digital twins, and real-time monitoring. Neural operators address this bottleneck by learning maps between function spaces, allowing a trained surrogate to approximate an entire solution operator rather than a single finite-dimensional regression map \citep{kovachki2023neuraloperator}. Architectures such as Fourier neural operators (FNOs), DeepONets, and convolutional neural operators (CNOs) have therefore become standard tools for PDE surrogate modeling \citep{li2021fourier,lu2021deeponet,raonic2023convolutional}.

As neural operators move toward deployable scientific workflows, the relevant bottleneck is not only predictive fidelity but also model size, memory footprint, throughput, and hardware availability. Full-size monolithic neural operators can be costly to store and evaluate, especially for high-resolution or three-dimensional fields. This has motivated parameter-efficient operator-learning methods based on factorization, tensorization, multigrid structure, mixed precision, pruning, quantization, and related compression mechanisms \citep{tran2023factorized,kossaifi2024multigrid,tu2024guaranteed,han2016deepcompressioncompressingdeep,novikov2015tensorizingNeuralNetworks,lu2025tensorcompressedfullyquantizedtrainingneural}. These constraints are especially important for edge and low-SWaP scientific systems, where surrogates may need to run near sensors, actuators, or embedded controllers rather than on centralized GPU servers \citep{howes2026realtimesensinginaccessiblephysical}.

This work studies a complementary route to parameter efficiency. Instead of training one large operator and then compressing it, we ask whether compact surrogates can be built directly as stagewise additive stacks of small neural operators. The idea is motivated by classical boosting, where predictors are constructed sequentially by fitting each new learner to the residual error of the current ensemble. Functional-gradient interpretations of boosting formalize this as greedy optimization in function space, and squared-loss boosting reduces naturally to residual fitting \citep{mason1999boostingGradientDescent,friedman2001gradientBoostingMachine,buhlmann2003boostingL2Loss}. For PDE surrogates, this viewpoint suggests that early small operators may capture dominant coarse structure in the solution map, while later residual operators focus capacity on systematic errors left by earlier stages, such as fine-scale features, spectral errors, boundary effects \citep{shikhman2026one}, or architecture-specific deficiencies. Recent spectral analyses of FNOs similarly indicate that residual correction can recover information missed by an initial operator \citep{qin2024betterunderstandingfourierneural}.

We introduce \emph{Operator Boosting}, a model-family-agnostic framework for constructing parameter-efficient PDE surrogates from existing neural-operator architectures. In the experiments, the framework is applied separately within each architecture family: a full FNO is compared against a boosted stack of tiny FNOs, a full DeepONet against a boosted stack of tiny DeepONets, and a full CNO against a boosted stack of tiny CNOs. This within-family protocol isolates the effect of stagewise residual learning from the effect of changing the operator architecture. The central question is whether boosted stacks can improve the deployment-relevant accuracy--parameter Pareto frontier relative to full-size monolithic baselines. The results should not be read as showing that residual boosting universally improves neural operators. Rather, they show that boosting is a simple wrapper that often improves the empirical accuracy--parameter tradeoff, while its failures reveal when the residual structure is poorly matched to the available tiny operator class.

The contributions of this work are:
\begin{itemize}
    \item We formulate \emph{Operator Boosting} as a stagewise functional-gradient framework for PDE surrogate modeling, where tiny neural operators are trained sequentially on residual fields and added with validation-selected shrinkage.
    \item We instantiate the framework with FNO, DeepONet, and CNO base learners, using within-family comparisons to separate the effect of the boosting procedure from the effect of changing architecture class.
    \item We evaluate boosted tiny stacks against full-size monolithic baselines across one-, two-, and three-dimensional PDE benchmarks from PDEBench, APEBench, and The Well \citep{takamoto2022pdebench,koehler2024apebench,ohana2024thewell}.
    \item We report both predictive error and trainable parameter count, emphasizing empirical accuracy--parameter Pareto tradeoffs rather than relative $L^2$ error alone.
    \item We identify Pareto-improving regimes, where boosted stacks achieve comparable or lower error with substantially fewer parameters, and failure regimes, where residual boosting exposes PDE- or architecture-dependent compression limits.
\end{itemize}

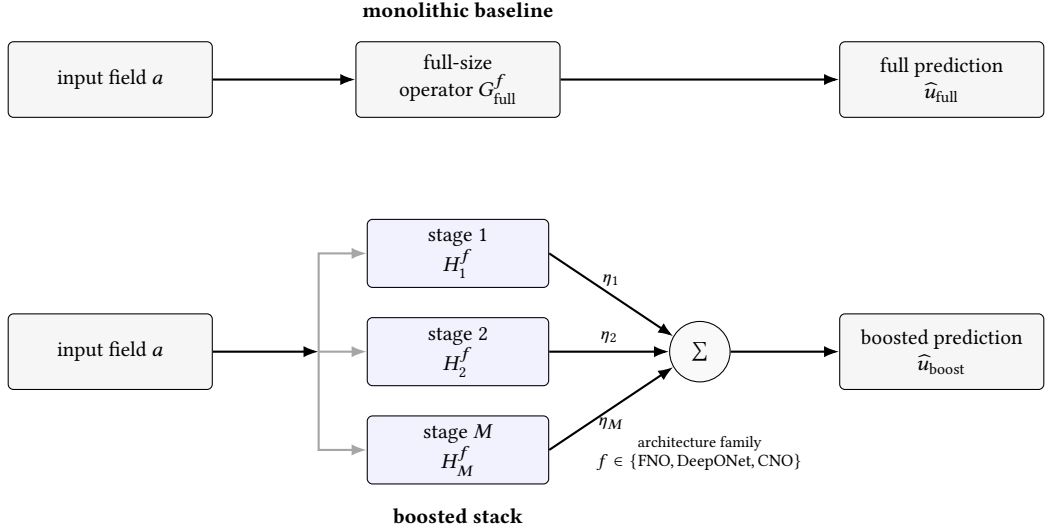
\begin{figure*}[t]
\centering
\begin{tikzpicture}[
    font=\small,
    box/.style={
        draw,
        rounded corners=2pt,
        minimum width=27mm,
        minimum height=10mm,
        align=center,
        fill=gray!7
    },
    tinybox/.style={
        draw,
        rounded corners=2pt,
        minimum width=24mm,
        minimum height=9mm,
        align=center,
        fill=blue!5
    },
    sum/.style={
        draw,
        circle,
        inner sep=1.5pt,
        minimum size=8mm,
        fill=gray!5
    },
    arrow/.style={-{Latex[length=2mm]}, thick},
    faintarrow/.style={-{Latex[length=2mm]}, thick, gray!70}
]

\coordinate (L) at (0,0);
\coordinate (C1) at (46mm,0);
\coordinate (C2) at (78mm,0);
\coordinate (R) at (110mm,0);

\coordinate (Lb) at (0,-36mm);
\coordinate (C1b) at (46mm,-36mm);
\coordinate (C2b) at (78mm,-36mm);
\coordinate (Rb) at (110mm,-36mm);

\node[box] (ain) at (L) {input field $a$};
\node[box] (full) at (C1) {full-size\\operator $G_{\mathrm{full}}^f$};
\node[box] (ufull) at (R) {full prediction\\$\widehat{u}_{\mathrm{full}}$};

\draw[arrow] (ain.east) -- (full.west);
\draw[arrow] (full.east) -- (ufull.west);

\node[above=2mm of full] {\textbf{monolithic baseline}};

\node[box] (ain2) at (Lb) {input field $a$};

\node[tinybox] (h1) at ($(C1b)+(0,13mm)$) {stage 1\\$H_1^f$};
\node[tinybox] (h2) at (C1b) {stage 2\\$H_2^f$};
\node[tinybox] (h3) at ($(C1b)+(0,-13mm)$) {stage $M$\\$H_M^f$};

\node[sum] (plus) at (C2b) {$\sum$};
\node[box] (uboost) at (Rb) {boosted prediction\\$\widehat{u}_{\mathrm{boost}}$};

\coordinate (split) at ($(ain2.east)+(14mm,0)$);
\draw[arrow] (ain2.east) -- (split);
\draw[faintarrow] (split) |- (h1.west);
\draw[faintarrow] (split) -- (h2.west);
\draw[faintarrow] (split) |- (h3.west);

\draw[arrow] (h1.east) -- node[above, font=\scriptsize] {$\eta_1$} (plus.150);
\draw[arrow] (h2.east) -- node[above, font=\scriptsize] {$\eta_2$} (plus.west);
\draw[arrow] (h3.east) -- node[below, font=\scriptsize] {$\eta_M$} (plus.210);

\draw[arrow] (plus.east) -- (uboost.west);

\node[below=2mm of h3] {\textbf{boosted stack}};

\node[below=6mm of plus, align=center, font=\scriptsize]
{architecture family\\$f \in \{\mathrm{FNO},\mathrm{DeepONet},\mathrm{CNO}\}$};

\end{tikzpicture}
\caption{Operator Boosting replaces a full-size monolithic neural operator with an additive stack of residual operators. Here $a$ denotes the input field, $G_0(a)=0$ is the normalized-output mean initializer, and each stage contributes a validation-shrunk correction $\eta_m H_m^f(a)$ to form the boosted prediction.}
\label{fig:operator_boosting_stack}
\end{figure*}

Figure~\ref{fig:operator_boosting_stack} demonstrates the difference between a full-size monolithic neural operator and the boosted residual stack used in Operator Boosting.


\section{Related Work}

\subsection{Neural operators for PDE surrogate modeling}

Neural operators learn maps between function spaces and have become a standard approach for PDE surrogate modeling, where the goal is to approximate solution operators mapping coefficients, initial conditions, forcing terms, or previous states to solution fields \citep{kovachki2023neuraloperator}. Early graph-based formulations include graph kernel and multipole graph neural operators \citep{anandkumar2019neural,li2020multipole}. Among widely used architectures, DeepONet represents operators through branch and trunk networks \citep{lu2021deeponet}, while Fourier neural operators learn global spectral updates on regular grids \citep{li2021fourier} and have been extended to more general geometries through learned deformations \citep{li2023geofno}. Convolutional neural operators provide a multiscale convolutional alternative for robust PDE learning \citep{raonic2023convolutional}. Benchmark suites such as PDEBench have further standardized empirical evaluation across PDE families \citep{takamoto2022pdebench}. In this work, FNO, DeepONet, and CNO serve as representative operator families for studying whether stagewise residual boosting can improve the accuracy--parameter Pareto frontier of PDE surrogates under severe parameter constraints.

\subsection{Ensembles, boosting, and residual learning}

Ensemble methods have long been used to improve predictive accuracy and stability by combining multiple learned predictors. Early neural-network ensembles showed that aggregating independently trained models can reduce generalization error \citep{hansen1990neural}, while bagging introduced bootstrap aggregation as a general-purpose variance-reduction strategy \citep{breiman1996bagging}. Boosting provides a more structured form of ensembling: rather than training predictors independently, it builds an additive model stage by stage, with later learners correcting the errors of earlier ones. This idea originates in the AdaBoost literature \citep{freund1997decisionTheoreticBoosting} and was later interpreted as gradient descent in function space \citep{mason1999boostingGradientDescent}. Friedman's gradient boosting framework formalized boosting as greedy functional approximation by fitting weak learners to pseudo-residuals \citep{friedman2001gradientBoostingMachine}, with stochastic variants improving robustness and scalability \citep{friedman2002stochasticGradientBoosting}. For squared-error regression, boosting with the $L_2$ loss is especially close to residual fitting, since each stage is trained to approximate the remaining prediction error \citep{buhlmann2003boostingL2Loss}. Modern deep learning has also made extensive use of ensemble and residual principles: deep ensembles remain a simple and effective approach for improving predictive reliability and uncertainty estimation \citep{lakshminarayanan2017deepEnsembles}, while residual networks showed that learning additive corrections can make very deep models easier to optimize \citep{he2016deepResidualLearning}. Operator Boosting transfers this residual-fitting principle to neural-operator surrogate modeling, where each weak learner is itself an operator mapping input fields to output fields.

\subsection{Parameter-efficient scientific machine learning}

Parameter efficiency has long been central to scientific computing, where high-fidelity PDE and dynamical-system solvers are often too expensive for repeated evaluation, uncertainty quantification, control, or design optimization. Classical reduced-order modeling addresses this issue by projecting high-dimensional dynamics onto compact approximation spaces \citep{benner2015projectionModelReduction}, while reduced-basis methods provide certified low-dimensional surrogates for parametrized PDEs and related field problems \citep{chen2010certifiedReducedBasisMaxwell}. In deep learning, parameter-efficient modeling has been studied through several complementary mechanisms, including knowledge distillation from large models or ensembles into smaller students \citep{hinton2015distillingknowledgeneuralnetwork}, pruning and quantization pipelines for compressed neural networks \citep{han2016deepcompressioncompressingdeep}, tensor-train representations of neural-network layers \citep{novikov2015tensorizingNeuralNetworks}, and sparse trainable subnetworks \citep{frankle2018the}. More recent scientific machine-learning work has adapted these ideas to neural PDE solvers and neural operators. Factorized Fourier neural operators reduce the cost of spectral operator learning by factorizing the Fourier representation \citep{tran2023factorized}, while multi-grid tensorized Fourier neural operators combine domain decomposition and tensorized parameter representations for high-resolution PDE surrogates \citep{kossaifi2024multigrid}. Mixed-precision neural operators target memory and throughput bottlenecks in FNO-style models while preserving approximation guarantees \citep{tu2024guaranteed}, and tensor-compressed, fully quantized neural PDE solvers further illustrate the growing interest in deployment-oriented scientific machine learning \citep{lu2025tensorcompressedfullyquantizedtrainingneural}. The present work is complementary to these compression and efficiency strategies. Rather than compressing a trained full-size operator, changing numerical precision, or imposing a low-rank parameterization on a single monolithic architecture, we study whether a stagewise additive stack of tiny neural operators can improve the empirical accuracy--parameter Pareto frontier directly.

\section{Problem Setting}

\subsection{PDE surrogate modeling}

Let $\Omega\subset\mathbb{R}^d$ denote the spatial domain and let $a$ denote the input field specifying the instance of a PDE surrogate problem. Depending on the benchmark, $a$ may represent an initial condition, a previous solution state, a coefficient field, a forcing field, or a collection of such quantities. The target field is denoted by $u$. The objective is to learn an operator
\begin{equation}
    \mathcal{G}:\mathcal{A}\to\mathcal{U},
    \qquad
    a \mapsto u,
\end{equation}
from paired observations
\begin{equation}
    \mathcal{D}=\{(a_i,u_i)\}_{i=1}^N.
\end{equation}
For static problems such as Darcy flow, $a$ is a coefficient or permeability field and $u$ is the corresponding solution field. For time-dependent problems, we construct one-step or finite-time transition pairs of the form
\begin{equation}
    a_i = \bigl(u(t_i),\Delta t_i\bigr),
    \qquad
    u_i = u(t_i+\Delta t_i),
\end{equation}
so that the learned operator approximates a time-advancement map. All models are trained and evaluated on discretized fields, but the surrogate-learning objective is operator-valued: the learned map should approximate the underlying input--output relation between fields rather than a scalar response.

\subsection{Full-size baseline and boosted-stack comparison}

For each dataset and architecture family $f$, we compare two predictors. The first is a full-size monolithic neural operator,
\begin{equation}
    G_{\mathrm{full}}^f:\mathcal{A}\to\mathcal{U},
\end{equation}
trained directly to predict the normalized output field. The second is a boosted stack of tiny same-family neural operators,
\begin{equation}
    G_{\mathrm{boost}}^f(a)
    =
    G_0(a)+\sum_{m=1}^M\eta_m H_m^f(a),
\end{equation}
where each $H_m^f$ is a small neural operator from the same architecture family as the full baseline and $\eta_m$ is a validation-selected shrinkage coefficient. In our experiments, $G_0(a)=0$ in normalized output coordinates, corresponding after denormalization to the empirical mean output field. Thus the boosted predictor contains no learned initialization model; its trainable parameters are entirely contained in the stagewise correction operators $H_m^f$.

This comparison is performed within each architecture family. A full FNO is compared against boosted tiny FNOs, a full DeepONet against boosted tiny DeepONets, and a full CNO against boosted tiny CNOs. This design isolates the effect of stagewise residual learning from the effect of switching model classes. The total boosted parameter count is
\begin{equation}
    P_{\mathrm{boost}}^f
    =
    \sum_{m=1}^M P(H_m^f),
\end{equation}
and, when all stages use the same tiny architecture,
\begin{equation}
    P_{\mathrm{boost}}^f = M P_{\mathrm{tiny}}^f.
\end{equation}
The full-size baseline parameter count is denoted by $P_{\mathrm{full}}^f$.

\subsection{Evaluation metrics and empirical Pareto dominance}

The primary accuracy metric is test relative $L^2$ error. For a predicted field $\widehat{u}$ and target field $u$, we compute
\begin{equation}
    \mathrm{RelL2}(\widehat{u},u)
    =
    \frac{\|\widehat{u}-u\|_2}{\|u\|_2+\varepsilon},
\end{equation}
where $\varepsilon>0$ is a small numerical stabilizer. Relative errors are computed after denormalizing predictions back to physical coordinates and are averaged over the test set.

For each dataset, architecture family, and random seed, we define the performance delta
\begin{equation}
    \Delta_{\mathrm{perf}}
    =
    100
    \frac{\mathrm{RelL2}_{\mathrm{full}}-\mathrm{RelL2}_{\mathrm{boost}}}{\mathrm{RelL2}_{\mathrm{full}}}.
\end{equation}
Thus $\Delta_{\mathrm{perf}}>0$ indicates that the boosted tiny stack outperforms the full-size monolithic baseline, while $\Delta_{\mathrm{perf}}<0$ indicates degradation. We also report the size delta
\begin{equation}
    \Delta_{\mathrm{size}}
    =
    100
    \frac{P_{\mathrm{boost}}-P_{\mathrm{full}}}{P_{\mathrm{full}}},
\end{equation}
so that negative values indicate parameter reduction. Equivalently, the size reduction is $-\Delta_{\mathrm{size}}$.

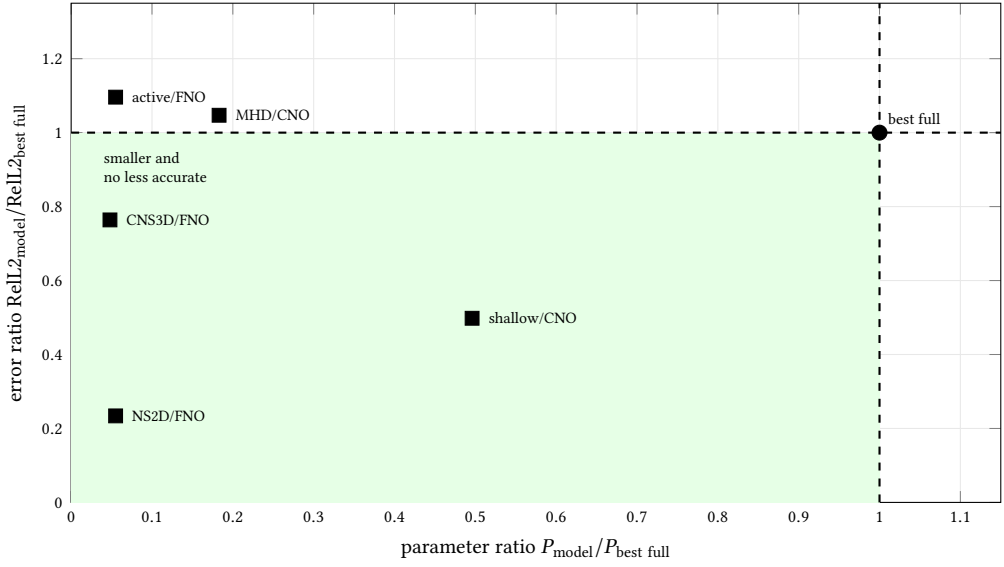
\begin{figure*}[t]
\centering
\begin{tikzpicture}
\begin{axis}[
    width=0.78\textwidth,
    height=0.46\textwidth,
    xmin=0, xmax=1.15,
    ymin=0, ymax=1.35,
    xlabel={parameter ratio $P_{\mathrm{model}}/P_{\mathrm{best\ full}}$},
    ylabel={error ratio $\mathrm{RelL2}_{\mathrm{model}}/\mathrm{RelL2}_{\mathrm{best\ full}}$},
    grid=both,
    grid style={gray!18},
    ticklabel style={font=\scriptsize},
    label style={font=\small},
    legend pos=north west,
    legend style={
        font=\scriptsize,
        draw=none,
        fill=white,
        fill opacity=0,
        text opacity=1,
        cells={anchor=west}
    },
    clip=false
]

\addplot[draw=none, fill=green!10] coordinates {(0,0) (1,0) (1,1) (0,1)} -- cycle;

\addplot[only marks, mark=*, mark size=2.8pt] coordinates {(1,1)};
\node[anchor=south west, font=\scriptsize] at (axis cs:1,1) {best full};

\addplot[only marks, mark=square*, mark size=2.6pt] coordinates {
    (0.055,0.234)  
    (0.496,0.498)  
    (0.048,0.764)  
    (0.055,1.096)  
    (0.183,1.047)  
};

\node[anchor=west, font=\scriptsize] at (axis cs:0.065,0.234) {NS2D/FNO};
\node[anchor=west, font=\scriptsize] at (axis cs:0.506,0.498) {shallow/CNO};
\node[anchor=west, font=\scriptsize] at (axis cs:0.058,0.764) {CNS3D/FNO};
\node[anchor=west, font=\scriptsize] at (axis cs:0.065,1.096) {active/FNO};
\node[anchor=west, font=\scriptsize] at (axis cs:0.193,1.047) {MHD/CNO};

\draw[dashed, thick] (axis cs:1,0) -- (axis cs:1,1.35);
\draw[dashed, thick] (axis cs:0,1) -- (axis cs:1.15,1);
\node[anchor=north west, font=\scriptsize, align=left] at (axis cs:0.03,0.97) {smaller and\\no less accurate};

\end{axis}
\end{tikzpicture}
\caption{Accuracy--parameter Pareto plane normalized by the best full-size baseline for each PDE. Points below and to the left of $(1,1)$ empirically Pareto-dominate the best full-size baseline. Representative boosted stacks show both strong Pareto improvements and compression tradeoffs.}
\label{fig:pareto_plane}
\end{figure*}

We say that a boosted stack empirically Pareto-dominates a full-size baseline when it achieves no larger mean test relative $L^2$ error while using fewer trainable parameters. That is, model $A$ dominates model $B$ if
\begin{equation}
    \overline{\mathrm{RelL2}}(A)\leq \overline{\mathrm{RelL2}}(B),
    \qquad
    P_A<P_B,
\end{equation}
with at least one strict improvement. In finite-seed experiments, we report accuracy and parameter-count changes separately rather than collapsing them into a scalar score. This distinguishes strong Pareto improvements, where the boosted stack is both smaller and more accurate, from compression tradeoffs, where parameter count is reduced at the cost of some predictive degradation. Figure~\ref{fig:pareto_plane} illustrates this normalized accuracy--parameter Pareto plane and shows representative dominating and tradeoff cases. 


\section{Operator Boosting}

\subsection{Stagewise additive operator model}

Let $\mathcal{A}$ denote the input function space and $\mathcal{U}$ the output function space. Given training data
\begin{equation}
    \mathcal{D}_{\mathrm{tr}}
    =
    \{(a_i,u_i)\}_{i=1}^{N_{\mathrm{tr}}},
    \qquad
    a_i\in\mathcal{A},\quad u_i\in\mathcal{U},
\end{equation}
the goal is to approximate an operator $\mathcal{G}:\mathcal{A}\to\mathcal{U}$. We work in normalized output coordinates. Let $\mu_y$ and $\sigma_y$ denote the empirical mean and standard deviation of the training outputs, computed over the training split, and define
\begin{equation}
    \widetilde{u}_i = \frac{u_i-\mu_y}{\sigma_y}.
\end{equation}
Operator Boosting represents the surrogate as an additive operator-valued model
\begin{equation}
    G_M(a)=G_0(a)+\sum_{m=1}^M\eta_m H_m(a),
\end{equation}
where each $H_m:\mathcal{A}\to\mathcal{U}$ is a small neural operator and $\eta_m\geq 0$ is a scalar shrinkage parameter. In this work, $G_0(a)=0$ in normalized coordinates, corresponding to the empirical mean predictor $\mu_y$ after denormalization. Thus the physical prediction is
\begin{equation}
    \widehat{u}_M(a)=\mu_y+\sigma_y G_M(a).
\end{equation}
This construction is a functional-gradient analogue of classical boosting, where an additive predictor is built by sequentially fitting weak learners to residual or pseudo-residual targets \citep{mason1999boostingGradientDescent,friedman2001gradientBoostingMachine,buhlmann2003boostingL2Loss}. Here the weak learners are neural operators mapping input fields to output fields.

\subsection{Residual fitting and shrinkage selection}

For squared-error loss in normalized coordinates, the negative functional gradient at stage $m$ is the residual field
\begin{equation}
    r_i^{(m)} = \widetilde{u}_i-G_{m-1}(a_i).
\end{equation}
For a fixed operator family $f$, the next correction is trained in the tiny model class $\mathcal{H}_{f,\mathrm{tiny}}$:
\begin{equation}
    H_m^f
    \in
    \arg\min_{H\in\mathcal{H}_{f,\mathrm{tiny}}}
    \frac{1}{N_{\mathrm{tr}}}
    \sum_{i=1}^{N_{\mathrm{tr}}}
    \left\|H(a_i)-r_i^{(m)}\right\|_2^2
    +
    \beta\,\mathcal{R}_{\mathrm{spec}}
    \!\left(H(a_i)-r_i^{(m)}\right),
\end{equation}
where $\mathcal{R}_{\mathrm{spec}}$ is an optional Fourier-weighted residual penalty. After training $H_m^f$, the shrinkage coefficient is selected on the validation set:
\begin{equation}
    \eta_m
    \in
    \arg\min_{\eta\in\Lambda}
    \frac{1}{N_{\mathrm{val}}}
    \sum_{j=1}^{N_{\mathrm{val}}}
    \left\|
        G_{m-1}(a_j^{\mathrm{val}})
        +
        \eta H_m^f(a_j^{\mathrm{val}})
        -
        \widetilde{u}_j^{\mathrm{val}}
    \right\|_2^2,
\end{equation}
where $\Lambda$ is a finite candidate grid containing $\eta=0$. Including $\eta=0$ makes each stage validation-safe: if a trained correction does not reduce validation error, it can be rejected. This validation-selected shrinkage plays a regularizing role analogous to shrinkage and early stopping in classical boosting \citep{zhang2005boostingEarlyStopping}. The boosted surrogate is then updated by
\begin{equation}
    G_m(a)=G_{m-1}(a)+\eta_m H_m^f(a).
\end{equation}
All primary reported results use the final $M$-stage predictor rather than selecting the best test-stage post hoc.

\subsection{Algorithm}

Algorithmic details are standard for squared-loss residual boosting: compute normalized residual fields, train a tiny correction operator for each stage, select shrinkage on the validation split, and add the accepted correction to the running surrogate. Full pseudo-code is provided in Appendix~\ref{app:pseudocode}.

\subsection{Instantiation across model classes}

The framework is model-family agnostic. We instantiate it with FNO, DeepONet, and CNO. Fourier neural operators use spectral convolution layers to learn global updates in Fourier space \citep{li2021fourier}; DeepONets use a branch--trunk decomposition motivated by universal approximation of nonlinear operators \citep{lu2021deeponet}; and convolutional neural operators use multiscale convolutional structure for operator learning on spatial fields \citep{raonic2023convolutional}. In experiments, all $H_m^f$ within a stack use the same tiny architecture, so
\begin{equation}
    P_{\mathrm{boost}}^f = M P_{\mathrm{tiny}}^f.
\end{equation}
Since $G_0$ is the zero predictor in normalized coordinates, it contributes no trainable parameters.

\section{Experimental Design}

\subsection{Datasets}

We evaluate Operator Boosting on a collection of one-, two-, and three-dimensional PDE surrogate-modeling tasks. The one-dimensional benchmarks are standardized advection, Burgers, and reaction--diffusion datasets. The two-dimensional benchmarks include Darcy flow and reaction--diffusion from PDEBench \citep{takamoto2022pdebench}, incompressible Navier--Stokes from APEBench \citep{koehler2024apebench}, shallow-water dynamics from PDEBench, and active matter from The Well \citep{ohana2024thewell}. The three-dimensional benchmarks are compressible Navier--Stokes from PDEBench and magnetohydrodynamics from The Well. For standardized datasets, each example is represented as a direct input--output pair $(a,u)$. For trajectory datasets, we construct supervised operator-learning pairs by sampling states separated by a time offset $\Delta t$, using the earlier state and normalized time offset as input and the later state as target. All datasets are normalized using statistics computed only from the training split.

The benchmark collection covers several mechanics-relevant surrogate regimes: transport-dominated dynamics in advection and Burgers problems, reaction--diffusion dynamics, elliptic porous-media-type prediction through Darcy flow, incompressible and compressible fluid dynamics through Navier--Stokes benchmarks, free-surface geophysical flow through shallow-water dynamics, and magnetohydrodynamic evolution. This range is intended to test whether the boosting mechanism is specific to simple low-dimensional PDEs or persists in higher-dimensional flow and field-evolution settings.

\subsection{Baselines}

For each PDE and architecture family, we compare a full-size monolithic neural operator against a boosted stack of tiny same-family operators. The comparisons are full FNO versus boosted tiny FNOs, full DeepONet versus boosted tiny DeepONets, and full CNO versus boosted tiny CNOs. This protocol separates the effect of the boosting procedure from the effect of changing architecture class. We also report best-model Pareto comparisons in which, for each PDE, the best full-size architecture is compared with the best boosted tiny stack among the completed runs. Parameter counts are computed directly from trainable model parameters. The present study focuses on full-size monolithic baselines because the primary question is whether stagewise stacks of tiny operators can improve the deployment-relevant accuracy--parameter frontier relative to standard full-size neural operators. Parameter-matched tiny monolithic models and independently trained tiny ensembles are natural additional baselines for future ablation.

\subsection{Training and aggregation summary}

All models use the same train/validation/test splits within each dataset. Full-size baselines are trained as monolithic operators, while boosted models use three tiny same-family residual stages with validation-selected shrinkage. Reported test errors are computed after denormalization to physical coordinates and aggregated over random seeds using seed-level means and 95\% confidence intervals. Full training and aggregation details are given in Appendix~\ref{app:reproducibility}.

\section{Results}

\subsection{Within-family accuracy--parameter tradeoffs}

Across all completed experiments, Operator Boosting improves the mean relative error in 21 of 30 dataset--architecture pairs, with 17 pairs having confidence intervals whose lower endpoint is positive. All boosted stacks reduce trainable parameter count, with mean reductions ranging from approximately 72\% to 95\%. The strongest gains occur for FNO on two-dimensional Navier--Stokes, where the boosted stack reduces mean relative error by 74.8\% while using 94.5\% fewer parameters, and for CNO and DeepONet on shallow-water dynamics, which obtain 51.9\% and 45.2\% reductions in relative error with 74.3\% and 88.1\% fewer parameters, respectively. Three-dimensional results further support the Pareto-efficiency interpretation: DeepONet improves on compressible Navier--Stokes by 27.6\% with 72.2\% fewer parameters, while FNO improves on MHD by 4.4\% with 95.2\% fewer parameters. At the same time, the failures are not isolated numerical noise: several PDE--architecture pairs exhibit negative mean gains, including CNO on Navier--Stokes, FNO on two-dimensional reaction--diffusion, and CNO on MHD. These failures motivate reporting the full accuracy--parameter tradeoff rather than only the successful Pareto-dominating cases. 

Table~\ref{tab:within_family_tradeoffs} reports the primary within-family comparison. Positive $\Delta_{\mathrm{perf}}$ means that the boosted tiny stack improves over the full-size baseline.

\begin{table*}[t]
\centering
\scriptsize
\setlength{\tabcolsep}{4pt}
\caption{Within-family accuracy--parameter tradeoffs grouped by architecture family. Positive $\Delta_{\mathrm{perf}}$ indicates lower test relative $L^2$ error for the boosted tiny stack than for the full-size baseline.}
\label{tab:within_family_tradeoffs}
\resizebox{\textwidth}{!}{%
\begin{tabular}{lcccrcc}
\toprule
Dataset & $n$ & Wins & RelL2 full $\to$ boosted & $\Delta_{\mathrm{perf}}$ (\%) & 95\% CI & Size red. (\%) \\
\midrule
\multicolumn{7}{l}{\textbf{FNO}} \\
\midrule
1D advection & 10 & 10/10 & $0.0592 \to 0.0305$ & 44.2 & [31.2, 57.2] & 86.2 \\
1D Burgers & 10 & 4/10 & $0.1095 \to 0.0967$ & -2.0 & [-21.0, 17.0] & 86.2 \\
1D reaction--diffusion & 10 & 10/10 & $7.580{\times}10^{-3} \to 4.074{\times}10^{-3}$ & 43.3 & [33.1, 53.5] & 86.2 \\
2D Navier--Stokes & 10 & 10/10 & $0.0805 \to 0.0188$ & 74.8 & [69.2, 80.4] & 94.5 \\
2D Darcy & 10 & 6/10 & $0.1383 \to 0.1334$ & 3.2 & [-2.3, 8.7] & 94.5 \\
2D reaction--diffusion & 10 & 3/10 & $0.0132 \to 0.0147$ & -20.6 & [-43.4, 2.3] & 94.5 \\
2D shallow water & 10 & 9/10 & $1.921{\times}10^{-3} \to 1.475{\times}10^{-3}$ & 19.9 & [6.6, 33.2] & 94.5 \\
2D active matter & 10 & 1/10 & $0.6575 \to 0.7206$ & -11.1 & [-24.0, 1.9] & 94.5 \\
3D compressible Navier--Stokes & 10 & 7/10 & $0.1790 \to 0.1368$ & 17.3 & [-1.8, 36.5] & 95.2 \\
3D MHD & 10 & 10/10 & $0.5104 \to 0.4879$ & 4.4 & [4.1, 4.7] & 95.2 \\
\midrule
\multicolumn{7}{l}{\textbf{DeepONet}} \\
\midrule
1D advection & 10 & 10/10 & $0.5009 \to 0.4127$ & 17.3 & [11.9, 22.8] & 87.6 \\
1D Burgers & 10 & 10/10 & $0.3620 \to 0.2806$ & 22.1 & [18.3, 25.9] & 87.6 \\
1D reaction--diffusion & 10 & 10/10 & $0.0942 \to 0.0735$ & 22.0 & [17.9, 26.1] & 87.6 \\
2D Navier--Stokes & 10 & 7/10 & $1.0366 \to 0.9853$ & 4.6 & [0.1, 9.2] & 84.5 \\
2D Darcy & 10 & 10/10 & $0.9133 \to 0.7683$ & 15.6 & [11.0, 20.2] & 84.5 \\
2D reaction--diffusion & 10 & 3/10 & $0.3087 \to 0.3095$ & -0.3 & [-0.8, 0.3] & 84.5 \\
2D shallow water & 10 & 10/10 & $0.0107 \to 5.787{\times}10^{-3}$ & 45.2 & [39.9, 50.5] & 88.1 \\
2D active matter & 10 & 7/10 & $1.1564 \to 1.1396$ & 1.3 & [-7.2, 9.8] & 87.9 \\
3D compressible Navier--Stokes & 10 & 7/10 & $0.3110 \to 0.2102$ & 27.6 & [2.3, 52.8] & 72.2 \\
3D MHD & 10 & 7/10 & $0.9452 \to 0.9018$ & 4.3 & [0.4, 8.3] & 72.3 \\
\midrule
\multicolumn{7}{l}{\textbf{CNO}} \\
\midrule
1D advection & 10 & 9/10 & $0.1657 \to 0.1380$ & 16.0 & [6.9, 25.1] & 74.5 \\
1D Burgers & 10 & 7/10 & $0.1510 \to 0.1230$ & 11.8 & [-5.0, 28.5] & 74.5 \\
1D reaction--diffusion & 10 & 10/10 & $0.0133 \to 0.0105$ & 21.0 & [13.1, 28.9] & 74.5 \\
2D Navier--Stokes & 10 & 0/10 & $0.0822 \to 0.2109$ & -202.0 & [-427.1, 23.1] & 74.3 \\
2D Darcy & 10 & 2/10 & $0.2079 \to 0.2260$ & -9.1 & [-16.0, -2.1] & 74.3 \\
2D reaction--diffusion & 10 & 8/10 & $0.0258 \to 0.0250$ & 2.9 & [0.1, 5.7] & 74.3 \\
2D shallow water & 10 & 10/10 & $2.215{\times}10^{-3} \to 9.572{\times}10^{-4}$ & 51.9 & [41.0, 62.9] & 74.3 \\
2D active matter & 10 & 3/10 & $0.7705 \to 0.7910$ & -6.3 & [-19.3, 6.6] & 74.3 \\
3D compressible Navier--Stokes & 10 & 4/10 & $0.2002 \to 0.2050$ & -18.8 & [-52.4, 14.9] & 81.7 \\
3D MHD & 10 & 0/10 & $0.4034 \to 0.4225$ & -4.8 & [-6.8, -2.8] & 81.7 \\
\bottomrule
\end{tabular}%
}
\end{table*}

\subsection{Best-model Pareto comparison}

At the dataset level, the best compact boosted surrogate empirically Pareto-dominates the best full-size monolithic surrogate on 7 of 10 completed PDE benchmarks. Dominating cases include one-dimensional advection, Burgers, and reaction--diffusion; two-dimensional Navier--Stokes, Darcy flow, and shallow-water dynamics; and three-dimensional compressible Navier--Stokes. The three tradeoff cases are two-dimensional reaction--diffusion, active matter, and MHD, where the boosted stack remains substantially smaller but does not match the lowest full-size baseline error. Detailed best-model tables are included in Appendix~\ref{app:additional_tables}.

\subsection{Failure modes}

The failures in Table~\ref{tab:within_family_tradeoffs} are consistent with prior diagnostic evidence that learned PDE surrogates exhibit strongly PDE- and architecture-dependent error modes~\citep{shikhman2026diagnosing}. Operator Boosting is least effective when the tiny correction operators cannot represent the relevant residual structure, when the full-size baseline already captures the dominant dynamics, or when residual fitting amplifies architecture-specific weaknesses. The clearest degradations occur for CNO on two-dimensional Navier--Stokes, FNO on two-dimensional reaction--diffusion, CNO on Darcy flow, CNO on MHD, and active matter for FNO and CNO. These cases indicate that Operator Boosting should be viewed as an accuracy--parameter Pareto-improvement strategy rather than a universal replacement for full-size neural operators. Its effectiveness depends jointly on the PDE family, the architecture class, and the residual structure left by earlier stages.

\section{Discussion}

\subsection{When does boosting help?}

The results suggest that Operator Boosting helps most when the full-size monolithic baseline is not simply parameter-limited, but when small residual operators can correct systematic errors left by earlier low-capacity stages. The strongest gains occur when the full model's error appears to contain a systematic component that remains representable by the tiny operator class. This is most evident for FNO on two-dimensional Navier--Stokes, where the boosted stack improves error by 74.8\% while using 94.5\% fewer parameters, and for shallow-water dynamics, where both CNO and DeepONet boosted stacks substantially improve over their full-size counterparts. These cases suggest that the stagewise residual objective is not merely acting as compression; it changes the optimization problem by allowing successive small operators to focus on the remaining error of the current surrogate. In contrast, failures such as CNO on Navier--Stokes and FNO on two-dimensional reaction--diffusion suggest that residual fitting is not sufficient when the tiny correction class poorly matches the residual geometry or when the residual is dominated by errors that require capacity absent from the tiny architecture.

\subsection{Implications for resource-constrained surrogate simulation}

The main deployment implication is that trainable parameter count can often be reduced by an order of magnitude without sacrificing accuracy, and in many cases while improving it. Parameter count is not identical to inference latency or peak memory, but it is a relevant proxy for storage, model transfer, and deployability. Because all correction operators receive the same input field at inference, their forward passes are independent after training and can in principle be evaluated in parallel. However, actual latency depends on hardware utilization, batching strategy, memory bandwidth, and the cost structure of the operator family. Thus the present results should be interpreted primarily as parameter-count and storage improvements rather than measured latency improvements.

\section{Limitations and Ethical Considerations}

This study focuses on trainable parameter count and relative $L^2$ error. These metrics capture the deployed accuracy--parameter tradeoff, but they do not fully measure deployment cost. Future evaluations should also report model size, peak activation memory, wall-clock latency, throughput, and energy use, especially because boosted stacks contain multiple forward modules even when their total parameter count is small. The experiments use fixed tiny architectures and three boosting stages; we do not optimize the number of stages, per-stage capacity, shrinkage grid, or residual loss weighting for each PDE. The method also remains architecture- and PDE-dependent: some PDE--family pairs degrade when the tiny correction class cannot represent the residual structure left by earlier stages.

The present work uses public scientific benchmark data and does not involve human subjects, private records, personally identifiable information, or intervention on physical systems. The main ethical risk is misuse of compact surrogates as replacements for validated high-fidelity solvers in safety-critical scientific or engineering decisions. Operator Boosting should therefore be used with domain validation, uncertainty assessment, and problem-specific physical diagnostics before deployment in settings where model error could affect safety, environmental outcomes, or infrastructure decisions.

\section{Conclusion}

This work introduced Operator Boosting, a stagewise residual-learning framework for constructing compact neural-operator PDE surrogates. Rather than training a large monolithic operator and compressing it afterward, Operator Boosting builds the surrogate directly as an additive stack of tiny same-family operators trained on successive residual fields, with each correction added through validation-selected shrinkage.

Across FNO, DeepONet, and CNO backbones, the results show that this residual construction can substantially improve the empirical accuracy--parameter tradeoff. Across 30 dataset--architecture pairs, 21 show positive mean accuracy gains, 17 have positive confidence intervals, and all boosted stacks reduce trainable parameter count by approximately 72--95\%. In best-model comparisons, boosted stacks empirically Pareto-dominate the best full-size baseline on 7 of 10 completed PDE benchmarks, including transport, reaction--diffusion, Darcy flow, shallow-water dynamics, and incompressible and compressible Navier--Stokes.

The experiments also show that Operator Boosting is not a universal replacement for full-size neural operators: several PDE--architecture pairs degrade despite large parameter reductions, indicating that success depends on whether the remaining residual is representable by the tiny correction class. Overall, the study demonstrates that residual stacks can improve the deployed size--accuracy frontier of neural PDE surrogates, while motivating future work on latency, memory footprint, energy use, stronger compression baselines, and adaptive choices of residual-stage size and structure.

\newpage

\section*{Acknowledgements}

\paragraph{Reproducibility} Code to reproduce all experiments, generate figures, and compute degradation metrics is available at \url{https://github.com/lennonshikhman/boosted-neural-operator}.

\paragraph{Computational Resources} The authors gratefully acknowledge Dell Technologies, and in particular the Dell Pro Precision division, for providing computational resources that supported the experiments in this work. All experiments were conducted on a Dell Pro Max T2 workstation equipped with an Intel Core Ultra 9 285K processor, 128 GB of DDR5 ECC memory, and an NVIDIA RTX PRO 6000 Blackwell GPU.

\section*{Generative AI Usage}

Generative AI tools were used only for formatting during manuscript preparation. The research ideas, experimental design, implementation, numerical results, analysis, and final scientific claims were checked and controlled by the author.

\bibliographystyle{ACM-Reference-Format}
\bibliography{references}

@misc{howes2026realtimesensinginaccessiblephysical,
      title={Real-Time Sensing of Inaccessible Physical Fields via an Edge-Deployable Hardware-Portable Graph Neural Operator}, 
      author={William Howes and Jason Yoo and Kazuma Kobayashi and Subhankar Sarkar and Farid Ahmed and Souvik Chakraborty and Syed Bahauddin Alam},
      year={2026},
      eprint={2604.01802},
      archivePrefix={arXiv},
      primaryClass={cs.LG},
      url={https://arxiv.org/abs/2604.01802}, 
}

@misc{qin2024betterunderstandingfourierneural,
      title={Toward a Better Understanding of Fourier Neural Operators from a Spectral Perspective}, 
      author={Shaoxiang Qin and Fuyuan Lyu and Wenhui Peng and Dingyang Geng and Ju Wang and Xing Tang and Sylvie Leroyer and Naiping Gao and Xue Liu and Liangzhu Leon Wang},
      year={2024},
      eprint={2404.07200},
      archivePrefix={arXiv},
      primaryClass={cs.LG},
      url={https://arxiv.org/abs/2404.07200}, 
}

@inproceedings{koehler2024apebench,
    title={{APEB}ench: A Benchmark for Autoregressive Neural Emulators of {PDE}s},
    author={Felix Koehler and Simon Niedermayr and r{\"u}diger westermann and Nils Thuerey},
    booktitle={The Thirty-eight Conference on Neural Information Processing Systems Datasets and Benchmarks Track},
    year={2024},
    url={https://openreview.net/forum?id=iWc0qE116u}
}

@article{zhang2005boostingEarlyStopping,
    author  = {Zhang, Tong and Yu, Bin},
    title   = {Boosting with Early Stopping: Convergence and Consistency},
    journal = {The Annals of Statistics},
    volume  = {33},
    number  = {4},
    pages   = {1538--1579},
    year    = {2005},
    month   = aug,
    doi     = {10.1214/009053605000000255},
    url     = {https://doi.org/10.1214/009053605000000255}
}

@inproceedings{ohana2024thewell,
 author = {Ohana, Ruben and McCabe, Michael and Meyer, Lucas and Morel, Rudy and Agocs, Fruzsina J. and Beneitez, Miguel and Berger, Marsha and Burkhart, Blakesley and Dalziel, Stuart B. and Fielding, Drummond B. and Fortunato, Daniel and Goldberg, Jared A. and Hirashima, Keiya and Jiang, Yan-Fei and Kerswell, Rich R. and Maddu, Suryanarayana and Miller, Jonah and Mukhopadhyay, Payel and Nixon, Stefan S. and Shen, Jeff and Watteaux, Romain and Blancard, Bruno R\'{e}galdo-Saint and Rozet, Fran\c{c}ois and Parker, Liam and Cranmer, Miles and Ho, Shirley},
 booktitle = {Advances in Neural Information Processing Systems},
 doi = {10.52202/079017-1430},
 editor = {A. Globerson and L. Mackey and D. Belgrave and A. Fan and U. Paquet and J. Tomczak and C. Zhang},
 pages = {44989--45037},
 publisher = {Curran Associates, Inc.},
 title = {The Well: a Large-Scale Collection of Diverse Physics Simulations for Machine Learning},
 url = {https://proceedings.neurips.cc/paper_files/paper/2024/file/4f9a5acd91ac76569f2fe291b1f4772b-Paper-Datasets_and_Benchmarks_Track.pdf},
 volume = {37},
 year = {2024}
}

@inproceedings{raonic2023convolutional,
    title={Convolutional Neural Operators for robust and accurate learning of {PDE}s},
    author={Bogdan Raonic and Roberto Molinaro and Tim De Ryck and Tobias Rohner and Francesca Bartolucci and Rima Alaifari and Siddhartha Mishra and Emmanuel de Bezenac},
    booktitle={Thirty-seventh Conference on Neural Information Processing Systems},
    year={2023},
    url={https://openreview.net/forum?id=MtekhXRP4h}
}

@article{lu2021deeponet,
  title   = {Learning nonlinear operators via {DeepONet} based on the universal approximation theorem of operators},
  author  = {Lu, Lu and Jin, Pengzhan and Pang, Guofei and Zhang, Zhongqiang and Karniadakis, George Em},
  journal = {Nature Machine Intelligence},
  volume  = {3},
  pages   = {218--229},
  year    = {2021},
  doi     = {10.1038/s42256-021-00302-5},
  url     = {https://doi.org/10.1038/s42256-021-00302-5}
}

@article{shikhman2026diagnosing,
    title={Diagnosing Failure Modes of Neural Operators Across Diverse {PDE} Families},
    author={Lennon Shikhman},
    journal={Transactions on Machine Learning Research},
    issn={2835-8856},
    year={2026},
    url={https://openreview.net/forum?id=0S1LWZHQYn},
    note={}
}

@inproceedings{shikhman2026one,
    title={One Operator to Rule Them All? On Boundary-Indexed Operator Families in Neural {PDE} Solvers},
    author={Lennon Shikhman},
    booktitle={AI{\&}PDE: ICLR 2026 Workshop on AI and Partial Differential Equations},
    year={2026},
    url={https://openreview.net/forum?id=lDjWQ9UxRy}
}

@inproceedings{li2021fourier,
    title={Fourier Neural Operator for Parametric Partial Differential Equations},
    author={Zongyi Li and Nikola Borislavov Kovachki and Kamyar Azizzadenesheli and Burigede liu and Kaushik Bhattacharya and Andrew Stuart and Anima Anandkumar},
    booktitle={International Conference on Learning Representations},
    year={2021},
    url={https://openreview.net/forum?id=c8P9NQVtmnO}
}

@article{kovachki2023neuraloperator,
  author  = {Nikola Kovachki and Zongyi Li and Burigede Liu and Kamyar Azizzadenesheli and Kaushik Bhattacharya and Andrew Stuart and Anima Anandkumar},
  title   = {Neural Operator: Learning Maps Between Function Spaces With Applications to PDEs},
  journal = {Journal of Machine Learning Research},
  year    = {2023},
  volume  = {24},
  number  = {89},
  pages   = {1--97},
  url     = {http://jmlr.org/papers/v24/21-1524.html}
}

@inproceedings{anandkumar2019neural,
    title={Neural Operator: Graph Kernel Network for Partial Differential Equations},
    author={Anima Anandkumar and Kamyar Azizzadenesheli and Kaushik Bhattacharya and Nikola Kovachki and Zongyi Li and Burigede Liu and Andrew Stuart},
    booktitle={ICLR 2020 Workshop on Integration of Deep Neural Models and Differential Equations},
    year={2019},
    url={https://openreview.net/forum?id=fg2ZFmXFO3}
}

@inproceedings{li2020multipole,
    author = {Li, Zongyi and Kovachki, Nikola and Azizzadenesheli, Kamyar and Liu, Burigede and Stuart, Andrew and Bhattacharya, Kaushik and Anandkumar, Anima},
    booktitle = {Advances in Neural Information Processing Systems},
    editor = {H. Larochelle and M. Ranzato and R. Hadsell and M.F. Balcan and H. Lin},
    pages = {6755--6766},
    publisher = {Curran Associates, Inc.},
    title = {Multipole Graph Neural Operator for Parametric Partial Differential Equations},
    url = {https://proceedings.neurips.cc/paper_files/paper/2020/file/4b21cf96d4cf612f239a6c322b10c8fe-Paper.pdf},
    volume = {33},
    year = {2020}
}

@inproceedings{takamoto2022pdebench,
    author = {Takamoto, Makoto and Praditia, Timothy and Leiteritz, Raphael and MacKinlay, Daniel and Alesiani, Francesco and Pfl\"{u}ger, Dirk and Niepert, Mathias},
    booktitle = {Advances in Neural Information Processing Systems},
    editor = {S. Koyejo and S. Mohamed and A. Agarwal and D. Belgrave and K. Cho and A. Oh},
    pages = {1596--1611},
    publisher = {Curran Associates, Inc.},
    title = {PDEBench: An Extensive Benchmark for Scientific Machine Learning},
    url = {https://proceedings.neurips.cc/paper_files/paper/2022/file/0a9747136d411fb83f0cf81820d44afb-Paper-Datasets_and_Benchmarks.pdf},
    volume = {35},
    year = {2022}
}

@article{li2023geofno,
  author  = {Zongyi Li and Daniel Zhengyu Huang and Burigede Liu and Anima Anandkumar},
  title   = {Fourier Neural Operator with Learned Deformations for PDEs on General Geometries},
  journal = {Journal of Machine Learning Research},
  year    = {2023},
  volume  = {24},
  number  = {388},
  pages   = {1--26},
  url     = {http://jmlr.org/papers/v24/23-0064.html}
}

@article{hansen1990neural,
    author  = {Hansen, L. K. and Salamon, P.},
    title   = {Neural Network Ensembles},
    journal = {IEEE Transactions on Pattern Analysis and Machine Intelligence},
    volume  = {12},
    number  = {10},
    pages   = {993--1001},
    year    = {1990},
    month   = oct,
    doi     = {10.1109/34.58871},
    url     = {https://doi.org/10.1109/34.58871}
}

@article{breiman1996bagging,
    author  = {Breiman, Leo},
    title   = {Bagging Predictors},
    journal = {Machine Learning},
    volume  = {24},
    pages   = {123--140},
    year    = {1996},
    doi     = {10.1007/BF00058655},
    url     = {https://doi.org/10.1007/BF00058655}
}

@article{freund1997decisionTheoreticBoosting,
    author  = {Freund, Yoav and Schapire, Robert E.},
    title   = {A Decision-Theoretic Generalization of On-Line Learning and an Application to Boosting},
    journal = {Journal of Computer and System Sciences},
    volume  = {55},
    number  = {1},
    pages   = {119--139},
    year    = {1997},
    doi     = {10.1006/jcss.1997.1504},
    url     = {https://doi.org/10.1006/jcss.1997.1504}
}

@inproceedings{mason1999boostingGradientDescent,
    author    = {Mason, Llew and Baxter, Jonathan and Bartlett, Peter and Frean, Marcus},
    title     = {Boosting Algorithms as Gradient Descent},
    booktitle = {Advances in Neural Information Processing Systems},
    volume    = {12},
    publisher = {MIT Press},
    year      = {1999},
    url       = {https://proceedings.neurips.cc/paper_files/paper/1999/file/96a93ba89a5b5c6c226e49b88973f46e-Paper.pdf}
}

@article{friedman2001gradientBoostingMachine,
    author  = {Friedman, Jerome H.},
    title   = {Greedy Function Approximation: A Gradient Boosting Machine},
    journal = {The Annals of Statistics},
    volume  = {29},
    number  = {5},
    pages   = {1189--1232},
    year    = {2001},
    month   = oct,
    doi     = {10.1214/aos/1013203451},
    url     = {https://doi.org/10.1214/aos/1013203451}
}

@article{friedman2002stochasticGradientBoosting,
    author  = {Friedman, Jerome H.},
    title   = {Stochastic Gradient Boosting},
    journal = {Computational Statistics \& Data Analysis},
    volume  = {38},
    number  = {4},
    pages   = {367--378},
    year    = {2002},
    doi     = {10.1016/S0167-9473(01)00065-2},
    url     = {https://doi.org/10.1016/S0167-9473(01)00065-2}
}

@article{buhlmann2003boostingL2Loss,
    author  = {B{\"u}hlmann, Peter and Yu, Bin},
    title   = {Boosting with the {$L_2$} Loss: Regression and Classification},
    journal = {Journal of the American Statistical Association},
    volume  = {98},
    number  = {462},
    pages   = {324--339},
    year    = {2003},
    doi     = {10.1198/016214503000125},
    url     = {https://doi.org/10.1198/016214503000125}
}

@inproceedings{lakshminarayanan2017deepEnsembles,
    author    = {Lakshminarayanan, Balaji and Pritzel, Alexander and Blundell, Charles},
    title     = {Simple and Scalable Predictive Uncertainty Estimation Using Deep Ensembles},
    booktitle = {Proceedings of the 31st International Conference on Neural Information Processing Systems},
    pages     = {6405--6416},
    year      = {2017},
    publisher = {Curran Associates Inc.}
}

@inproceedings{he2016deepResidualLearning,
    author    = {He, Kaiming and Zhang, Xiangyu and Ren, Shaoqing and Sun, Jian},
    title     = {Deep Residual Learning for Image Recognition},
    booktitle = {Proceedings of the IEEE Conference on Computer Vision and Pattern Recognition},
    pages     = {770--778},
    year      = {2016},
    doi       = {10.1109/CVPR.2016.90}
}

@article{benner2015projectionModelReduction,
    author  = {Benner, Peter and Gugercin, Serkan and Willcox, Karen},
    title   = {A Survey of Projection-Based Model Reduction Methods for Parametric Dynamical Systems},
    journal = {SIAM Review},
    volume  = {57},
    number  = {4},
    pages   = {483--531},
    year    = {2015},
    doi     = {10.1137/130932715},
    url     = {https://doi.org/10.1137/130932715}
}

@article{chen2010certifiedReducedBasisMaxwell,
    author  = {Chen, Yanlai and Hesthaven, Jan S. and Maday, Yvon and Rodr{'i}guez, Jer{'o}nimo},
    title   = {Certified Reduced Basis Methods and Output Bounds for the Harmonic {Maxwell}'s Equations},
    journal = {SIAM Journal on Scientific Computing},
    volume  = {32},
    number  = {2},
    pages   = {970--996},
    year    = {2010},
    doi     = {10.1137/09075250X},
    url     = {https://doi.org/10.1137/09075250X}
}

@misc{hinton2015distillingknowledgeneuralnetwork,
      title={Distilling the Knowledge in a Neural Network}, 
      author={Geoffrey Hinton and Oriol Vinyals and Jeff Dean},
      year={2015},
      eprint={1503.02531},
      archivePrefix={arXiv},
      primaryClass={stat.ML},
      url={https://arxiv.org/abs/1503.02531}, 
}

@misc{han2016deepcompressioncompressingdeep,
      title={Deep Compression: Compressing Deep Neural Networks with Pruning, Trained Quantization and Huffman Coding}, 
      author={Song Han and Huizi Mao and William J. Dally},
      year={2016},
      eprint={1510.00149},
      archivePrefix={arXiv},
      primaryClass={cs.CV},
      url={https://arxiv.org/abs/1510.00149}, 
}

@inproceedings{tran2023factorized,
    title={Factorized Fourier Neural Operators},
    author={Alasdair Tran and Alexander Mathews and Lexing Xie and Cheng Soon Ong},
    booktitle={The Eleventh International Conference on Learning Representations },
    year={2023},
    url={https://openreview.net/forum?id=tmIiMPl4IPa}
}

@article{kossaifi2024multigrid,
    title={Multi-Grid Tensorized Fourier Neural Operator for High- Resolution {PDE}s},
    author={Jean Kossaifi and Nikola Borislavov Kovachki and Kamyar Azizzadenesheli and Anima Anandkumar},
    journal={Transactions on Machine Learning Research},
    issn={2835-8856},
    year={2024},
    url={https://openreview.net/forum?id=AWiDlO63bH},
    note={}
}

@inproceedings{
    frankle2018the,
    title={The Lottery Ticket Hypothesis: Finding Sparse, Trainable Neural Networks},
    author={Jonathan Frankle and Michael Carbin},
    booktitle={International Conference on Learning Representations},
    year={2019},
    url={https://openreview.net/forum?id=rJl-b3RcF7},
}

@inproceedings{novikov2015tensorizingNeuralNetworks,
    author    = {Novikov, Alexander and Podoprikhin, Dmitrii and Osokin, Anton and Vetrov, Dmitry},
    title     = {Tensorizing Neural Networks},
    booktitle = {Advances in Neural Information Processing Systems},
    volume    = {28},
    publisher = {Curran Associates, Inc.},
    year      = {2015},
    url       = {https://proceedings.neurips.cc/paper_files/paper/2015/file/6855456e2fe46a9d49d3d3af4f57443d-Paper.pdf}
}

@inproceedings{tu2024guaranteed,
    title={Guaranteed Approximation Bounds for Mixed-Precision Neural Operators},
    author={Renbo Tu and Colin White and Jean Kossaifi and Boris Bonev and Gennady Pekhimenko and Kamyar Azizzadenesheli and Anima Anandkumar},
    booktitle={The Twelfth International Conference on Learning Representations},
    year={2024},
    url={https://openreview.net/forum?id=QJGj07PD9C}
}

@misc{lu2025tensorcompressedfullyquantizedtrainingneural,
      title={Tensor-Compressed and Fully-Quantized Training of Neural PDE Solvers}, 
      author={Jinming Lu and Jiayi Tian and Yequan Zhao and Hai Li and Zheng Zhang},
      year={2025},
      eprint={2512.09202},
      archivePrefix={arXiv},
      primaryClass={cs.LG},
      url={https://arxiv.org/abs/2512.09202}, 
}

\clearpage
\appendix

\section{Operator Boosting Pseudo-code}\label{app:pseudocode}

Algorithm~\ref{alg:operator_boosting} summarizes the training procedure for a fixed operator family $f$.

\begin{algorithm}[H]
\caption{Operator Boosting}
\label{alg:operator_boosting}
\begin{algorithmic}[1]
\Require Training set $\mathcal{D}_{\mathrm{tr}}$, validation set $\mathcal{D}_{\mathrm{val}}$, operator family $f$, number of stages $M$, shrinkage grid $\Lambda$
\State Compute output normalization statistics $\mu_y$ and $\sigma_y$ on $\mathcal{D}_{\mathrm{tr}}$
\State Normalize outputs by $\widetilde{u}=(u-\mu_y)/\sigma_y$
\State Initialize $G_0(a)=0$
\For{$m=1,\ldots,M$}
    \State Compute residual targets $r_i^{(m)}=\widetilde{u}_i-G_{m-1}(a_i)$ for all training samples
    \State Train a tiny neural operator $H_m^f\in\mathcal{H}_{f,\mathrm{tiny}}$ on $\{(a_i,r_i^{(m)})\}_{i=1}^{N_{\mathrm{tr}}}$
    \State Select
    \[
        \eta_m
        \in
        \arg\min_{\eta\in\Lambda}
        \sum_{(a_j,\widetilde{u}_j)\in\mathcal{D}_{\mathrm{val}}}
        \left\|
            G_{m-1}(a_j)+\eta H_m^f(a_j)-\widetilde{u}_j
        \right\|_2^2
    \]
    \State Update $G_m(a)=G_{m-1}(a)+\eta_m H_m^f(a)$
\EndFor
\State \Return Denormalized predictor $\widehat{u}_M(a)=\mu_y+\sigma_yG_M(a)$
\end{algorithmic}
\end{algorithm}

\section{Reproducibility Details}\label{app:reproducibility}

All models are trained under a fixed-budget terminal-iterate protocol. Full-size baselines are trained for a prescribed number of epochs and evaluated at the final epoch. Each boosted correction stage is also trained for the same prescribed stage budget and evaluated at its final epoch. At stage $m$, the current residual $r_i^{(m)}=\widetilde{u}_i-G_{m-1}(a_i)$ is used as the target for the next tiny operator. After training the correction operator, the shrinkage coefficient $\eta_m$ is selected on the validation set from a finite grid $\Lambda$ containing $0$. Unless otherwise stated, each boosted stack uses three stages. Full baselines are trained with mean-squared error, while boosted residual stages may include a small Fourier-weighted residual penalty to emphasize high-frequency residual structure. All reported relative errors are computed after denormalizing the predicted fields back to physical coordinates.

Operator Boosting reduces deployed parameter count, but it does not necessarily reduce training cost. Each boosted stack requires training $M$ residual operators, whereas each full-size baseline requires training a single monolithic model. The reported parameter counts therefore characterize deployed surrogate size rather than total training compute. This distinction is important in offline surrogate construction, where additional training cost may be acceptable if the resulting model is repeatedly deployed in many-query workflows.

\subsection{Statistical aggregation}

Results are aggregated over random seeds using seed-level means and 95\% confidence intervals. For each PDE and model family, we report the mean performance delta, confidence interval, number of wins, mean full-baseline error, mean boosted-stack error, and parameter reduction. In addition to within-family comparisons, we report best-model Pareto tables comparing the best full-size surrogate and the best boosted tiny surrogate for each PDE.

\section{Additional Aggregated Result Tables}\label{app:additional_tables}

The following tables provide the detailed best-model comparisons summarized in the main text.

\subsection{Best full baseline versus boosted stack}

Table~\ref{tab:best_full_same_family_boosted} selects, for each dataset, the full-size architecture with the lowest mean test relative $L^2$ error and compares it with the boosted stack from the same architecture family. This isolates whether boosting improves the strongest monolithic family for each PDE.

\begin{table*}[t]
\centering
\scriptsize
\setlength{\tabcolsep}{4pt}
\caption{Best full-size baseline versus its boosted tiny stack. For each dataset, the best full-size architecture is selected by mean test relative $L^2$ error and compared with the boosted tiny stack from the same family.}
\label{tab:best_full_same_family_boosted}
\resizebox{\textwidth}{!}{%
\begin{tabular}{llrrrrr}
\toprule
Dataset & Full family & Full RelL2 & Boosted RelL2 & $\Delta_{\mathrm{perf}}$ (\%) & Size red. (\%) & Wins \\
\midrule
1D advection & FNO & 0.0592 & 0.0305 & 44.2 & 86.2 & 10/10 \\
1D Burgers & FNO & 0.1095 & 0.0967 & -2.0 & 86.2 & 4/10 \\
1D reaction--diffusion & FNO & $7.580{\times}10^{-3}$ & $4.074{\times}10^{-3}$ & 43.3 & 86.2 & 10/10 \\
2D Navier--Stokes & FNO & 0.0805 & 0.0188 & 74.8 & 94.5 & 10/10 \\
2D Darcy & FNO & 0.1383 & 0.1334 & 3.2 & 94.5 & 6/10 \\
2D reaction--diffusion & FNO & 0.0132 & 0.0147 & -20.6 & 94.5 & 3/10 \\
2D shallow water & FNO & $1.921{\times}10^{-3}$ & $1.475{\times}10^{-3}$ & 19.9 & 94.5 & 9/10 \\
2D active matter & FNO & 0.6575 & 0.7206 & -11.1 & 94.5 & 1/10 \\
3D compressible Navier--Stokes & FNO & 0.1790 & 0.1368 & 17.3 & 95.2 & 7/10 \\
3D MHD & CNO & 0.4034 & 0.4225 & -4.8 & 81.7 & 0/10 \\
\bottomrule
\end{tabular}%
}
\end{table*}

The best full-size baseline is FNO for most completed datasets under the current experimental configuration, with CNO selected for three-dimensional MHD. Boosted FNOs substantially improve advection, reaction--diffusion, Navier--Stokes, shallow water, and three-dimensional compressible Navier--Stokes, give a small positive tradeoff on Darcy flow, and underperform on Burgers, two-dimensional reaction--diffusion, active matter, and MHD when compared against the best full-size family for those cases.

\subsection{Best full baseline versus best boosted stack}

Table~\ref{tab:best_full_vs_best_boosted} allows the boosted stack family to differ from the best full-size family. This comparison measures the deployment-relevant Pareto frontier: whether the best compact boosted surrogate can outperform the best full-size monolithic surrogate for the same PDE. Operator Boosting produces empirical Pareto improvements on 7 of 10 datasets. The three failures are active matter, two-dimensional reaction--diffusion, and MHD, where compression is obtained but the best boosted stack does not match the best full-size baseline error.

\begin{table*}[t]
\centering
\scriptsize
\setlength{\tabcolsep}{4pt}
\caption{Best full-size baseline versus best boosted tiny stack. For each dataset, the best full-size baseline and best boosted tiny surrogate are selected by mean test relative $L^2$ error. This comparison permits architecture selection among completed boosted stacks.}
\label{tab:best_full_vs_best_boosted}
\resizebox{\textwidth}{!}{%
\begin{tabular}{llrlrrrl}
\toprule
Dataset & Best full family & Full RelL2 & Best boosted family & Boosted RelL2 & RelL2 red. (\%) & Size red. (\%) & Pareto status \\
\midrule
1D advection & FNO & 0.0592 & FNO & 0.0305 & 48.6 & 86.2 & Dominates \\
1D Burgers & FNO & 0.1095 & FNO & 0.0967 & 11.6 & 86.2 & Dominates \\
1D reaction--diffusion & FNO & $7.580{\times}10^{-3}$ & FNO & $4.074{\times}10^{-3}$ & 46.3 & 86.2 & Dominates \\
2D Navier--Stokes & FNO & 0.0805 & FNO & 0.0188 & 76.6 & 94.5 & Dominates \\
2D Darcy & FNO & 0.1383 & FNO & 0.1334 & 3.5 & 94.5 & Dominates \\
2D reaction--diffusion & FNO & 0.0132 & FNO & 0.0147 & -11.1 & 94.5 & Tradeoff \\
2D shallow water & FNO & $1.921{\times}10^{-3}$ & CNO & $9.572{\times}10^{-4}$ & 50.2 & 50.4 & Dominates \\
2D active matter & FNO & 0.6575 & FNO & 0.7206 & -9.6 & 94.5 & Tradeoff \\
3D compressible Navier--Stokes & FNO & 0.1790 & FNO & 0.1368 & 23.6 & 95.2 & Dominates \\
3D MHD & CNO & 0.4034 & CNO & 0.4225 & -4.7 & 81.7 & Tradeoff \\
\bottomrule
\end{tabular}%
}
\end{table*}

\section{Artifact Availability}

The experimental code, data-loading scripts, aggregation scripts, and figure-generation source will be released upon acceptance. The benchmark datasets used in this study are publicly available through their respective sources, including PDEBench, APEBench, and The Well, together with the standardized synthetic PDE datasets used for the one-dimensional experiments.

\end{document}